\documentclass[lettersize,journal]{IEEEtran}
\usepackage{fontspec}        
\usepackage{polyglossia}    

\setdefaultlanguage{english}
\setotherlanguage{bengali}

\newfontfamily\bangla[
  Script=Bengali,
  Path=./fonts/,
  Extension=.ttf,
  UprightFont = *-Light,
  BoldFont = *-Bold
]{NotoSerifBengali}

\usepackage{amsmath, amssymb, amssymb, amsfonts}
\usepackage{algorithmic}
\usepackage{graphicx}
\usepackage{orcidlink}
\usepackage{enumitem}
\usepackage{textcomp}
\usepackage{xcolor}
\usepackage{url}
\usepackage{booktabs}
\usepackage{float}
\usepackage{array}
\usepackage{multirow}
\usepackage{xcolor,colortbl}
\usepackage{tikz}
\usepackage{underscore}
\usepackage{varwidth}
\usepackage{makecell}  
\usepackage{hyperref}

\usepackage{subcaption}

\usepackage{textcomp}
\usepackage{stfloats}
\usepackage{verbatim}

\usepackage{listings}
\usepackage[dvipsnames]{xcolor}

\definecolor{lightgreen}{RGB}{220,255,220}
\definecolor{lightred}{RGB}{220,0,0}

\definecolor{codegray}{rgb}{0.5,0.5,0.5}
\definecolor{lightash}{rgb}{0.98,0.98,0.98}

\usepackage[linesnumbered,ruled,vlined]{algorithm2e}
\usepackage{float}

\hypersetup{
	pdftitle={Retrieval Augmented Enhanced Dual Co-Attention Framework for Target Aware Multimodal Bengali Hateful Meme Detection},
	pdfauthor={Raihan Tanvir}
}

\usepackage[backend=biber,sorting=none]{biblatex}
\newcommand\submittedtext{%
  \footnotesize This work has been submitted to the IEEE Transactions on Computational Social Systems for  possiblepublication. 
  Copyright may be transferred without notice, after which this version may no longer be accessible.}

\newcommand\submittednotice{%
  \begin{tikzpicture}[remember picture,overlay]
    \node[anchor=south,yshift=8pt] at (current page.south) 
    {\fbox{\parbox{\dimexpr0.8\textwidth-\fboxsep-\fboxrule\relax}{\submittedtext}}};
  \end{tikzpicture}%
}

\begin{document}

\title{Retrieval Augmented Enhanced Dual Co-Attention Framework for Target Aware Multimodal Bengali Hateful Meme Detection}

\author{
Raihan Tanvir\,\orcidlink{0000-0003-3673-3401}
\and
Md. Golam Rabiul Alam\,\orcidlink{0000-0002-9054-7557}
\thanks{

Raihan Tanvir is with the Department of Computer Science and Engineering, BRAC University, Dhaka 1212, Bangladesh, and also with the Department of Computer Science and Engineering, Ahsanullah University of Science and Technology, Dhaka 1208, Bangladesh (e-mail: raihan.tanvir@g.bracu.ac.bd).

Md. Golam Rabiul Alam is with the Department of Computer Science and Engineering, BRAC University, Dhaka 1212, Bangladesh (e-mail: rabiul.alam@bracu.ac.bd).


}
}




\maketitle
\submittednotice

\begin{abstract}
Hateful content on social media often manifests as multimodal memes, blending images and text to propagate harmful messages. In low-resource languages like Bengali, detecting such memes is hindered by data scarcity, class imbalance, and complex code-mixing. To address these issues: 1) we augment the Bengali Hateful Memes (BHM) dataset with semantically aligned samples from the MultIMOdal aggreSsion dAtaset in Bengali (MIMOSA) to enhance class balance and diversity; 2) we propose the Enhanced Dual cO-attention fRAmework (xDORA), integrating vision encoders (\texttt{CLIP}, \texttt{DINOv2}) and multilingual text encoders (\texttt{XGLM}, \texttt{XLM-R}) with weighted attention pooling for robust crossmodal features; 3) we develop a Facebook AI Similarity Search (FAISS)-Based k-NN classifier using xDORA embeddings for non-parametric classification; 4) we introduce RAG-Fused DORA, leveraging FAISS-based retrieval for enhanced contextual understanding; and 5) we evaluate \texttt{LLaVA} in zero-shot, few-shot, and RAG-Prompted settings for low-resource Bengali contexts. Experiments on the extended dataset demonstrate that xDORA (\texttt{CLIP + XLM-R}) achieves macro-average F1-scores of 0.78 for hateful meme identification and 0.71 for target entity detection, with RAG-Fused DORA improving these to 0.79 and 0.74, yielding 3.95\% and 23.33\% gains over the DORA baseline (0.76 and 0.60, respectively). These models outperform baseline approaches, including several state-of-the-art unimodal and multimodal models, due to enhanced multimodal fusion and retrieval augmentation. The FAISS-Based k-NN approach performs competitively, excelling for rare classes by leveraging semantic proximity. In contrast, \texttt{LLaVA}'s few-shot performance (F1: 0.53 for meme identification, 0.26 for target detection) improves marginally in RAG-Prompted settings (0.54 and 0.39), highlighting the limitations of pretrained vision-language models for code-mixed Bengali content without fine-tuning. TThese results highlight the effectiveness of supervised, retrieval-augmented, and non-parametric multimodal frameworks in tackling linguistic and cultural complexities for hate speech detection in low-resource settings, laying the groundwork for future advancements in adaptable and extensible solutions.\\
\textcolor{red}{Disclaimer: This paper contains elements that one might find offensive which cannot be avoided due to the nature of the work.}
\end{abstract}


\section{Introduction}
\label{sec:introduction}

\IEEEPARstart{I}{nternet} memes, which integrate visual imagery and textual content, have become a significant medium for communication on social media platforms, frequently employed to express humor, convey opinions, or provide cultural commentary. However, these memes can also serve as vehicles for disseminating harmful content, targeting individuals, communities, organizations, or societal groups, thereby amplifying social divisions. In Bangladesh, the proliferation of hateful memes targeting minorities, religious leaders, and political figures has intensified social tensions, further complicated by cross-border trolling involving neighboring countries. The detection of such hateful memes is critical to mitigating their societal impact. Despite this urgency, research has predominantly focused on high-resource languages such as English, leaving low-resource languages like Bengali—spoken by over 230 million people worldwide \cite{alam2021bangla}—significantly underexplored.

\begin{figure}[ht]
	\centering
	\includegraphics[width=0.5\linewidth]{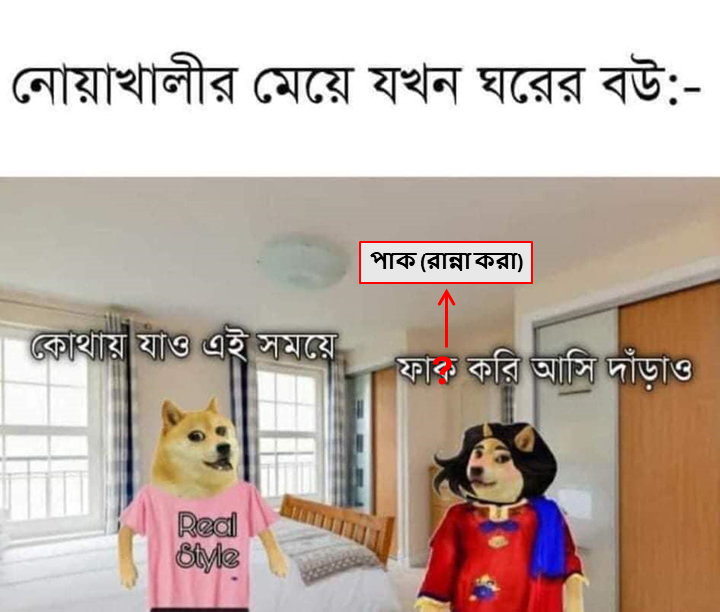}
	\caption{A sample meme from the Bengali Hateful Memes dataset, showcasing nuanced visual and textual elements that convey hateful sentiment.}
	\label{fig:sample_meme}
\end{figure}

Hossain et al. \cite{hossain2024deciphering} introduced the Bengali Hateful Memes (BHM) dataset, a groundbreaking multimodal resource designed for detecting hateful memes through binary (hateful vs. non-hateful) and multiclass (categorizing targets as individuals, communities, organizations, or societal groups) classification tasks. Their proposed Dual cO-attention fRAmework (DORA) leverages a co-attention mechanism to effectively integrate visual and textual representations, achieving superior performance compared to existing state-of-the-art baselines. Figure~\ref{fig:sample_meme} exemplifies a meme from the BHM dataset, where subtle visual and textual cues combine to disparage a specific regional population in Bangladesh, highlighting the complex interplay of multimodal content. Nevertheless, the BHM dataset is hindered by class imbalance, particularly in the multiclass task, which restricts the generalizability of trained models. To address this limitation, we enhance the BHM dataset by incorporating 2,233 samples from the MultIMOdal aggreSsion dAtaset in Bengali (MIMOSA) \cite{ahsan2024multimodal} through semantic label remapping, resulting in a more balanced and diverse dataset comprising 9,342 samples.

Building upon the DORA framework, we propose the Enhanced Dual cO-attention fRAmework (xDORA), which integrates a diverse set of vision encoders, including the vision encoder of Contrastive Language-Image Pre-training (CLIP) with Vision Transformer (ViT) \cite{radford2021learning} and Self-Distillation with No Labels -- Version 2 (DINOv2) \cite{oquab2023dinov2}, and multilingual text encoders, such as Cross-lingual Generative Language Model (XGLM) \cite{xglm} and Cross-lingual Language Model-RoBERTa (XLM-R) \cite{xlmroberta}, with weighted attention pooling to produce robust crossmodal representations. Additionally, we introduce a FAISS-Based k-Nearest Neighbor (k-NN) classifier utilizing xDORA embeddings for non-parametric classification. We also propose RAG-Fused DORA, an ensemble model that combines xDORA predictions with similarity-weighted probabilities derived from samples retrieved using Facebook AI Similarity Search (FAISS) \cite{johnson2019billion}, leveraging Retrieval-Augmented Generation (RAG) \cite{lewis2020rag} to enhance contextual understanding for imbalanced classes. Furthermore, we evaluate the large vision-language model Large Language and Vision Assistant (LLaVA) \cite{liu2023visualinstructiontuning} in zero-shot and few-shot settings, both independently and as RAG-Prompted LLaVA, to assess its efficacy in low-resource Bengali contexts.

This study makes three key contributions to Bengali hateful meme classification:
\begin{enumerate}
	\item We enhance the BHM dataset by integrating MIMOSA samples through semantic label remapping, improving dataset diversity and addressing class imbalance.
	\item We develop \textbf{xDORA}, a robust multimodal architecture that integrates state-of-the-art vision encoders (CLIP, DINOv2) and multilingual text encoders (XGLM, XLM-R) with weighted attention pooling for enhanced crossmodal fusion. Furthermore, we propose \textbf{RAG-Fused DORA}, a hybrid ensemble that combines supervised predictions with FAISS-based similarity-weighted label aggregation to improve detection accuracy and generalization, particularly for underrepresented classes.
	\item We conduct comprehensive evaluation of \textbf{LLaVA} under zero-shot, few-shot, and \textbf{RAG-Prompted} settings for Bengali multimodal content, revealing both its limitations and potential for retrieval-augmented adaptation.
\end{enumerate}

Our findings highlight the efficacy of retrieval-augmented approaches in addressing challenges related to data scarcity and class imbalance, laying the foundation for robust hateful content detection in low-resource languages. The paper is organized as follows: Section~\ref{sec:related_works} reviews related work, Section~\ref{sec:dataset} elaborates on dataset augmentation, Section~\ref{sec:methodology} details the proposed methodologies, Section~\ref{sec:experiments} presents experimental results and analysis, and Section~\ref{sec:conclusion} summarizes conclusions and outlines prospects for future research.

\section{Related Works}
    \label{sec:related_works}

    The proliferation of hateful content on social media, particularly through memes that combine visual and textual elements, has driven research into multimodal hate speech detection, leveraging vision-language models to capture the complex interplay between text and images. Kiela et al. \cite{kiela2021hateful} introduced the Hateful Memes dataset for English, demonstrating improved detection performance using multimodal transformers. However, their work focuses on high-resource languages, leaving low-resource languages like Bengali underexplored. Kumar and Nandakumar \cite{kumar2022hateclipper} proposed \textit{Hate-CLIPper}, which models cross-modal interactions using a Feature Interaction Matrix (FIM) over embeddings derived from the CLIP with Vision Transformer (ViT) \cite{dosovitskiy2020image}. This approach achieved a state-of-the-art AUROC score of 85.8 on the Hateful Memes Challenge, surpassing human performance. While effective in high-resource settings, such architectures have not been extensively applied to low-resource languages like Bengali, where code-mixing and cultural nuances present additional challenges.

    Recent advancements in multimodal hate speech detection have introduced innovative techniques to address both implicit and explicit hateful content in memes. Arya et al. \cite{arya2024multimodal} utilized the Contrastive Language-Image Pre-Training (CLIP) model, fine-tuned with prompt engineering, to achieve an accuracy of 87.42\% on the Facebook Hateful Memes dataset, with strong performance in AUROC and F1-score metrics. Wang et al. \cite{wang2025few} proposed a few-shot in-context learning framework that incorporates socio-cultural knowledge retrieval, metaphorical tenor identification, and chain-of-thought prompting with multimodal large language models, outperforming supervised methods on hateful meme benchmarks. Additionally, Zhang et al. \cite{zhang2025flexible} introduced Flexible Optimal Transport (FLOT) with contrastive graphical modeling to capture non-literal cross-modal alignments in implicit hateful memes, achieving state-of-the-art results across multiple benchmark datasets without relying on external knowledge.

    In the context of Bengali, research has focused on addressing the challenges of code-mixed and culturally nuanced content. Hossain et al. \cite{hossain2024align} developed a multimodal framework with context-aware attention mechanisms that dynamically align textual and visual data by prioritizing salient features. Evaluated on the MultiOFF (English) \cite{suryawanshi2020multimodal} and MUTE (Bengali code-mixed) \cite{hossain2022mute} datasets, their approach achieved F1-scores of 70.3\% and 69.7\%, respectively, demonstrating generalizability over existing state-of-the-art methods. Further advancing Bengali hateful meme detection, Hossain et al. \cite{hossain2024deciphering} introduced the BHM dataset and the Dual cO-attention fRAmework (DORA). The BHM dataset supports binary and multiclass classification, while DORA integrates CLIP (ViT) and XGLM \cite{xglm} with a co-attention mechanism, achieving robust performance. However, class imbalance in the BHM dataset, particularly for multiclass tasks, limits model generalizability, and DORA's reliance on standard attention mechanisms restricts advanced feature integration.

    Other efforts in Bengali meme analysis have tackled related challenges. Ahsan et al. \cite{ahsan2024multimodal} introduced the MIMOSA dataset, focusing on aggressive content targeting social entities, providing a valuable resource for studying harmful memes. Hossain et al. \cite{hossain2022mute, hossain2022memosen} developed datasets for hateful meme detection and sentiment analysis but did not address targeted entity classification or class imbalance. Karim et al. \cite{karim2022multimodalhate} explored hate speech in Bengali memes and texts, but their dataset is not publicly available and lacks focus on targeted entity classification. Das et al. \cite{das2023banglaabusememe} introduced BanglaAbuseMeme, a dataset of 4,043 annotated Bengali memes, yet it does not address challenges posed by cross-lingual captions.

    Large Vision-Language Models (LVLMs) offer promising avenues for low-resource applications. The Large Language and Vision Assistant (LLaVA) \cite{liu2023visualinstructiontuning} enables zero-shot reasoning by aligning CLIP (ViT) with language models, though its efficacy for code-mixed Bengali memes with cultural nuances remains untested. Retrieval-Augmented Generation (RAG) has emerged as a powerful approach to enhance hate speech detection by incorporating external knowledge. Sunkara et al. \cite{sunkara2024war} utilized RAG with large language models to classify, counter, and diffuse hate speech, achieving superior performance on a large tweet corpus. This approach underscores RAG’s potential to improve contextual understanding in low-resource settings like Bengali, where class imbalance and code-mixing pose challenges. Similarly, Zhang et al. \cite{zhang2024rezg} proposed ReZG, a RAG-based method for zero-shot counter-narrative generation, leveraging multi-dimensional hierarchical retrieval to address unseen targets, which is particularly relevant for Bengali hateful meme detection.

    Building on these works, our study enhances dataset diversity, model performance, and scalability for Bengali hateful meme classification. We address the limitations of class imbalance and low-resource contexts through an augmented BHM dataset, advanced multimodal architectures like the Enhanced Dual cO-attention fRAmework (xDORA) with state-of-the-art vision encoders such as DINOv2 \cite{oquab2023dinov2} and multilingual text encoders like XLM-R \cite{xlmroberta}, and retrieval-augmented techniques, including RAG-Fused DORA and RAG-Prompted LLaVA, supported by FAISS \cite{johnson2019billion}.

\section{Dataset}
\label{sec:dataset}

This study utilizes the Bengali Hateful Memes (BHM) dataset \cite{hossain2024deciphering} as the primary resource, augmented with selected samples from the MultIMOdal aggreSsion dAtaset in Bengali (MIMOSA) \cite{ahsan2024multimodal} to address class imbalance and enhance diversity. Below, we describe the original BHM dataset, including its composition and classification tasks, and the augmentation process, including the resulting extended dataset for both binary and multiclass classification tasks.

\subsection{BHM Dataset}

The BHM dataset, introduced by Hossain et al. \cite{hossain2024deciphering}, comprises 7,109 memes collected from public social media platforms in Bangladesh, such as Facebook and Instagram, targeting Bengali-speaking audiences. It supports two classification tasks: binary classification (Task 1: Hate vs. Non-Hate) and multiclass classification (Task 2: categorizing hateful memes by target). A meme is deemed hateful if it overtly seeks to demean, disparage, injure, ridicule, or harass an entity based on characteristics such as gender, ethnicity, ideology, religion, socioeconomic status, political alignment, regional identity, or organizational ties. For the multiclass task, hateful memes are categorized into four target classes: Targeted Individual (TI), targeting specific persons (e.g., celebrities like Sakib Khan or politicians like Khaleda Zia); Targeted Organization (TO), targeting groups with common goals (e.g., businesses like Grameenphone or political parties like Awami League); Targeted Community (TC), targeting groups sharing beliefs or ideologies (e.g., religious groups like Buddhists or cultural communities like Pohela Boishakh celebrants); and Targeted Society (TS), targeting populations based on geographical identity (e.g., stereotyping Indian or British communities).

Table~\ref{tab:bhm-stats} presents the dataset statistics across training, validation, and testing splits. In Task 1, the non-hate class significantly outweighs the hate class, with approximately 1.7 times more non-hate samples than hate samples, indicating a notable class imbalance. For Task 2, the TI class dominates, with roughly 6.5 times more samples than TC, nearly 10 times more than TO, and 19.7 times more than TS, highlighting severe class imbalance in the multiclass task.

\begin{table}[!htbp]
	\centering
	\caption{Number of memes in training, validation, and testing splits for the BHM dataset.}
	\label{tab:bhm-stats}
	\small
	\resizebox{0.9\columnwidth}{!}{%
		\begin{tabular}{@{}l|ccccc@{}}
			\toprule
			& \textbf{Class} & \textbf{Train} & \textbf{Valid} & \textbf{Test} & \textbf{Total} \\ 
			\midrule
			\multirow{2}{*}{Task 1} 
			& Hate  & 2117 & 241 & 266 & 2624 \\
			& Non-Hate & 3641 & 399 & 445 & 4485 \\
			\midrule
			\multirow{4}{*}{Task 2} 
			& TI & 1623 & 192 & 193 & 2008 \\
			& TO & 160  & 17  & 27  & 204 \\
			& TC & 249  & 24  & 37  & 310 \\
			& TS & 85   & 8   & 9   & 102 \\
			\bottomrule
		\end{tabular}%
	}
\end{table}

The memes in the BHM dictionary were annotated by multiple human annotators to ensure reliable labeling for both tasks. Inter-annotator agreement was measured using Cohen's \(\kappa\)-score, as shown in Table~\ref{tab:kappa-scores}. For Task 1, the average \(\kappa\)-score is 0.79, indicating substantial agreement among annotators. In contrast, Task 2 has a lower average \(\kappa\)-score of 0.63, reflecting moderate agreement and the increased complexity of identifying specific targets in hateful memes.

\begin{table}[!htb]
	\centering
	\caption{Cohen's \(\kappa\) agreement scores for BHM dataset annotations.}
	\label{tab:kappa-scores}
	\small
	\resizebox{0.7\columnwidth}{!}{%
		\begin{tabular}{@{}l|lcc@{}}
			\toprule
			& \textbf{Label} & \textbf{\(\kappa\)-score} & \textbf{Average} \\ 
			\midrule
			\multirow{2}{*}{Task 1} 
			& Hate & 0.82 & \multirow{2}{*}{0.79} \\
			& Non-Hate & 0.76 & \\ 
			\midrule
			\multirow{4}{*}{Task 2} 
			& TI & 0.68 & \multirow{4}{*}{0.63} \\
			& TO & 0.66 & \\
			& TC & 0.61 & \\
			& TS & 0.57 & \\ 
			\bottomrule
		\end{tabular}%
	}
\end{table}

\subsection{Dataset Augmentation}

To address the class imbalance in the BHM dataset and enhance its diversity, we augmented it with 2,233 carefully selected samples from the MIMOSA dataset \cite{ahsan2024multimodal}, which originally contains 4,848 samples across five target labels: Political, Gender, Religious, Others, and Non-aggression. After excluding 1,783 samples that overlap with the BHM dataset, 3,065 unique samples were considered for augmentation.

Since the MIMOSA dataset was curated for aggression detection, its labels do not directly align with the BHM dataset’s hate categories. We performed a taxonomy-driven label remapping to ensure compatibility. Specifically, Political, Gender, and Religious samples were mapped to the BHM hate categories of TO, TI, and TC respectively. A subset of Non-aggression samples was retained and assigned to the Non-Hate class, while the remainder were excluded due to semantic ambiguity. The ‘Others’ class was discarded, as it lacked a semantically aligned counterpart in the BHM taxonomy. This remapping was guided by the BHM dataset’s definitions of hateful content, and a random subset of the remapped data underwent manual review to ensure label integrity, maintaining consistency with the original hate classification framework.

The resulting extended BHM dataset comprises 9,342 samples for binary classification and 4,857 samples for multiclass classification. Table~\ref{tab:extended-stats} presents the statistics across training, validation, and testing splits, following an 80\%-10\%-10\% distribution with stratification. For Task 1, the hate to non-hate ratio improves to near parity, significantly reducing imbalance. For Task 2, the class distribution is more balanced, with TI samples approximately 2 times more than TC, 2.6 times more than TO, and 21.6 times more than TS. While the TS class remains underrepresented, the augmentation substantially increases the representation of TC and TO classes, enhancing the dataset’s robustness and diversity.

\begin{table}[!htb]
	\centering
	\caption{Number of memes in the extended BHM dataset.}
	\label{tab:extended-stats}
	\small
	\resizebox{0.9\columnwidth}{!}{%
		\begin{tabular}{@{}l|ccccc@{}}
			\toprule
			& \textbf{Class} & \textbf{Train} & \textbf{Valid} & \textbf{Test} & \textbf{Total} \\ 
			\midrule
			\multirow{2}{*}{Task 1} 
			& Hate  & 3885 & 486 & 486 & 4857 \\
			& Non-Hate & 3588 & 449 & 448 & 4485 \\
			\midrule
			\multirow{4}{*}{Task 2} 
			& TI & 2026 & 253 & 253 & 2532 \\
			& TC & 981  & 123 & 122 & 1226 \\
			& TO & 786  & 98  & 98  & 982 \\
			& TS & 94   & 12  & 11  & 117 \\
			\bottomrule
		\end{tabular}%
	}
\end{table}

\begin{figure*}[!t]
	\centering
	\includegraphics[width=1\textwidth]{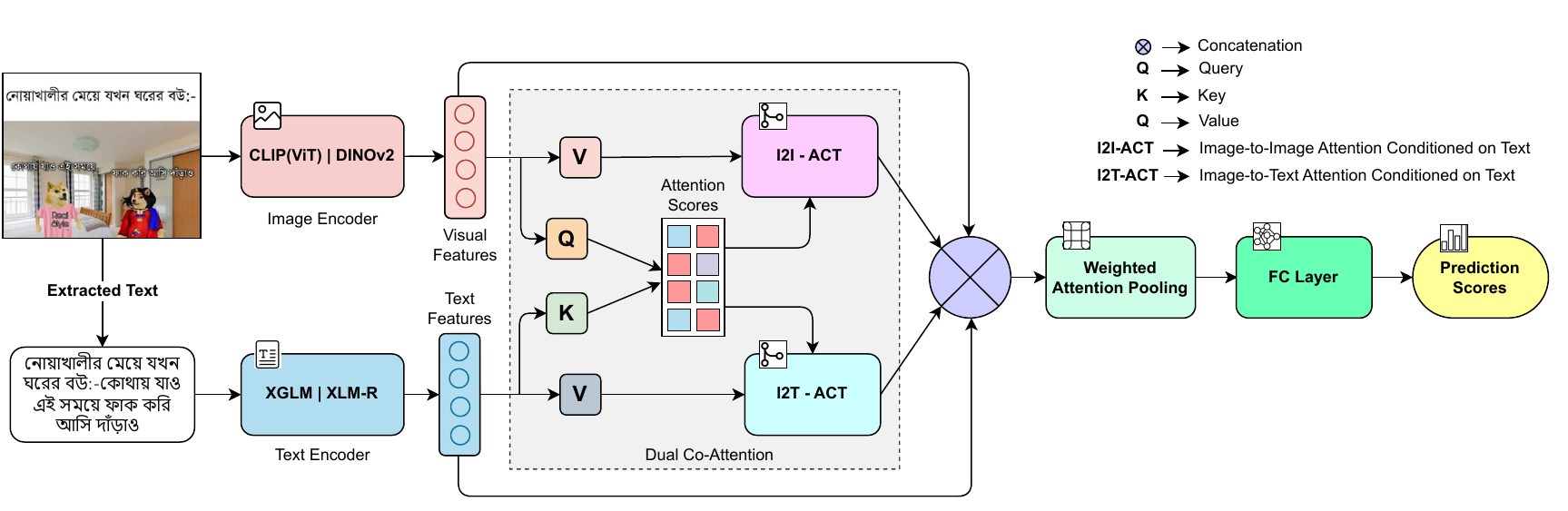}
	\caption{Overview of the xDORA architecture. Visual and textual features are extracted using pretrained encoders. A dual co-attention module with multi-head attention integrates the modalities through two cross-attention flows: I2T-ACT (Image-to-Text Attention Conditioned on Text, Equation~\eqref{eq:attn1}) and I2I-ACT (Image-to-Image Attention Conditioned on Text, Equation~\eqref{eq:attn2}). The fused representation is aggregated via weighted attention pooling and fed to a multilayer perceptron for classification.}
	\label{fig:xdora-architecture}
\end{figure*}

\section{Methodology}
\label{sec:methodology}

We propose four approaches for Bengali hateful meme classification: an Enhanced Dual cO-attention fRAmework (xDORA) for supervised learning, a FAISS-based non-parametric classifier, and two retrieval-augmented strategies: RAG-Fused DORA and RAG-Prompted LLaVA

\subsection{Enhanced DORA Architecture}
\label{subsec:method-xdora}

The Enhanced DORA (\textit{xDORA}) framework advances the original DORA model \cite{hossain2024deciphering} by integrating robust encoder pairs and a refined dual co-attention mechanism for high-fidelity vision-language fusion. We implement and integrate four encoder combinations: (1) \texttt{CLIP} (ViT-B/32) \cite{radford2021learning} with \texttt{XGLM-564M} \cite{xglm}, (2) \texttt{CLIP} (ViT-B/32) with \texttt{XLM-R-L} (XLM-RoBERTa-Large) \cite{xlmroberta}, (3) \texttt{DINOv2-Base} \cite{oquab2023dinov2} with \texttt{XGLM-564M}, and (4) \texttt{DINOv2-Base} with \texttt{XLM-R-L}. For vision processing, \texttt{CLIP} (ViT-B/32) employs a Vision Transformer with 32$\times$32 patches and contrastive learning for semantic alignment between image and text, while \texttt{DINOv2-Base} captures spatial and semantic features through self-distillation without labels. For text encoding, \texttt{XGLM-564M} is a multilingual autoregressive transformer with 564 million parameters designed for generative tasks in low-resource languages like Bengali, whereas \texttt{XLM-R-L} is a RoBERTa-based \cite{liu2019roberta} multilingual model pretrained on over 100 languages to provide deep contextual embeddings.

\subsubsection{Vision and Text Feature Extraction}

Let \(\mathbf{I}\) denote an input image, and \(\mathbf{X}_{\text{ids}}\) be the tokenized caption with corresponding attention mask \(\mathbf{X}_{\text{mask}}\). The visual encoders, \texttt{CLIP} (ViT-B/32) and \texttt{DINOv2-Base}, produce features with dimensions \(d_v = 512\) and \(d_v = 768\), respectively (Equation \eqref{eq:vision}). These are projected to match the text encoder's dimension \(d_t = 1024\) via a linear layer (Equation \eqref{eq:proj_vision}) and pooled to align with the text token sequence length \(S\) (Equation \eqref{eq:final_vision}).

\begin{subequations}
	\begin{align}
		\mathbf{V}_0 &= \text{VisualEncoder}(\mathbf{I}) \in \mathbb{R}^{d_v} 
		\label{eq:vision} \\
		\mathbf{V}_1 &= \mathbf{V}_0 \cdot \mathbf{W}_v + \mathbf{b}_v \in \mathbb{R}^{d_t}
		\label{eq:proj_vision} \\
		\mathbf{V} &= \text{AvgPool}(\mathbf{V}_1) \in \mathbb{R}^{S \times d_t} \label{eq:final_vision}
	\end{align}
\end{subequations}

The output from the text encoder, which processes the input tokens \(\mathbf{X}_{\text{ids}}\), produces contextualized embeddings with dimension \(d_t\) as shown in \eqref{eq:text}:

\begin{equation}
	\mathbf{T} = \text{TextEncoder}(\mathbf{X}_{\text{ids}}, \mathbf{X}_{\text{mask}}) \in \mathbb{R}^{S \times d_t} 
	\label{eq:text}
\end{equation}

\subsubsection{Dual Co-Attention Mechanism}

To facilitate rich cross-modal interaction, a dual co-attention mechanism is employed using multihead attention. The first attention map, termed \textbf{Image-to-Text Attention Conditioned on Text (I2T-ACT)} (Equation \eqref{eq:attn1}), uses visual features \(\mathbf{V}\) as the Query (\(Q\)) and text features \(\mathbf{T}\) as both Key (\(K\)) and Value (\(V\)). This allows visual tokens to selectively attend to relevant textual cues, capturing text-guided visual attention..

The second attention map, termed \textbf{Image-to-Image Attention Conditioned on Text (I2I-ACT)} (Equation \eqref{eq:attn2}), uses visual features \(\mathbf{V}\) as Query and Value, with text features \(\mathbf{T}\) as Key. This refines visual features by conditioning their self-attention on textual cues, preserving intra-modal visual relationships while enhancing multimodal alignment. This dual configuration enables robust vision-language integration by capturing both inter-modal and intra-modal dependencies.


\begin{subequations}
\begin{align}
	\mathbf{A}_1 &= \text{MultiheadAttn}(Q=\mathbf{V}, K=\mathbf{T}, V=\mathbf{T}) \in \mathbb{R}^{S \times d_t} \label{eq:attn1}\\
	\mathbf{A}_2 &= \text{MultiheadAttn}(Q=\mathbf{V}, K=\mathbf{T}, V=\mathbf{V}) \in \mathbb{R}^{S \times d_t} \label{eq:attn2}
\end{align}
\label{eq:attn}
\end{subequations}

\subsubsection{Fusion and Aggregation}

The attended outputs \(\mathbf{A}_1\) and \(\mathbf{A}_2\), along with the original visual and textual embeddings, are concatenated across the feature dimension to yield a fused representation \(\mathbf{F}\) (Equation \eqref{eq:concat}). To distill a single, unified vector representation, a soft attention mechanism is applied over the sequence dimension (Equation \eqref{eq:attn_w}). The token-level representations are weighted and aggregated to yield the pooled feature vector \(\mathbf{Z} \in \mathbb{R}^{4d_t} = \mathbb{R}^{4096}\), as expressed in \eqref{eq:fused_output}.

\begin{subequations}
	\begin{align}
		\mathbf{F} &= [\mathbf{A}_1 \| \mathbf{A}_2 \| \mathbf{V} \| \mathbf{T}] \in \mathbb{R}^{S \times 4d_t}
		\label{eq:concat} \\
		\boldsymbol{\alpha} &= \text{softmax}(\mathbf{F}, \text{dim}=1) 
		\label{eq:attn_w} \\
		\mathbf{Z} &= \sum_{s=1}^{S} \boldsymbol{\alpha}_s \odot \mathbf{F}_s \in \mathbb{R}^{4d_t} 
		\label{eq:fused_output}
	\end{align}
\end{subequations}

\subsubsection{Classification Layer}

The pooled embedding \(\mathbf{Z}\) is fed into a two-layer multilayer perceptron (MLP) classifier with LeakyReLU activation and dropout regularization. The output of the final linear layer yields the logits \(\mathbf{O}\) (Equation \eqref{eq:logit}), which are transformed into class probabilities \(\hat{\mathbf{y}}\) by applying an activation function as shown in \eqref{eq:xdora-proba}.

\begin{subequations}
	\begin{align}
		\mathbf{O} &= \text{MLP}(\mathbf{Z}) \in \mathbb{R}^C 
		\label{eq:logit} \\
		\hat{\mathbf{y}} &=
		\begin{cases}
			[1 - \sigma(\mathbf{O}),\; \sigma(\mathbf{O})], & \text{if } C = 2 \\[4pt]
			\text{softmax}(\mathbf{O}), & \text{if } C = 4
		\end{cases} \label{eq:xdora-proba}
	\end{align}
\end{subequations}

Here, \(C = 2\) for hateful meme detection and \(C = 4\) for targeted group identification.

\begin{figure*}[!htb]
	\centering
	\includegraphics[width=\linewidth]{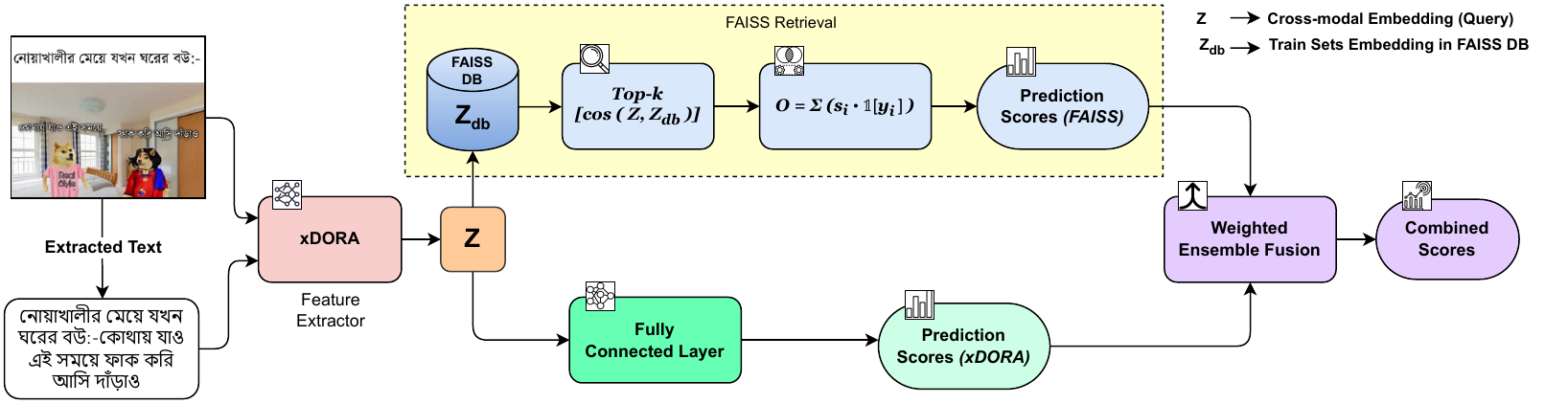}
	\caption{Architecture of RAG-Fused DORA. A multimodal embedding \(\mathbf{Z}\) is obtained using xDORA as a feature extractor. This embedding is used in two parallel branches: (1) it is passed through an MLP classifier to produce prediction scores; and (2) it is used to retrieve the top-\(k\) nearest neighbors from a FAISS index based on cosine similarity (\(\text{Top-}k[\cos(\mathbf{Z}, \mathbf{Z}_{\text{db}})]\)). The retrieved labels are aggregated using similarity-weighted label distribution. Final predictions are computed via a weighted ensemble of the classifier logits and the FAISS-derived label distribution.}
	\label{fig:rag-fused-dora}
\end{figure*}

\subsection{FAISS-Based k-Nearest Neighbor Classifier}
\label{subsec:method-knn}

The FAISS-based k-Nearest Neighbor (k-NN) classifier employs non-parametric classification through similarity search, leveraging multimodal embeddings to address class imbalance in the BHM dataset. The Facebook AI Similarity Search (FAISS) library enables efficient similarity search and clustering of high-dimensional vectors \cite{johnson2019billion}. The workflow, consisting of embedding generation, top-\(k\) retrieval, and similarity-weighted label aggregation, is integrated as a module within the RAG-Fused DORA architecture (see Fig.~\ref{fig:rag-fused-dora}).

\subsubsection{Embedding Generation}

For each training instance, a 4096-dimensional embedding is generated using the xDORA model's fused representation \(\mathbf{Z} \in \mathbb{R}^{4096}\) (Equation \ref{eq:fused_output}). Embeddings are normalized using the \(\ell_2\)-norm, producing \(\tilde{\mathbf{Z}} = \frac{\mathbf{Z}}{\|\mathbf{Z}\|_2} \in \mathbb{R}^{4096}\), where \(\|\mathbf{Z}\|_2 = \sqrt{\sum_{i=1}^{4096} Z_i^2}\) is the Euclidean norm. These normalized embeddings are stored in a FAISS FlatL2 index for efficient similarity search.

\subsubsection{Top-\textit{k} Retrieval}

For a test instance with normalized embedding \(\tilde{\mathbf{Z}}_{\text{test}}\), FAISS retrieves the top-\(k\) nearest neighbors from the training set, where \(k\) is a hyperparameter:

\begin{equation}
	\mathcal{N}(\tilde{\mathbf{Z}}_{\text{test}}) = \{(\tilde{\mathbf{Z}}_j, y_j, s_j)\}_{j=1}^k,
	\label{eq:knn_retrieval}
\end{equation}
where \(\tilde{\mathbf{Z}}_j \in \mathbb{R}^{4096}\) is the normalized embedding of the \(j\)-th neighbor, \(y_j \in \{0, 1, \ldots, C-1\}\) is its class label (\(C = 2\) for hateful meme detection, \(C = 4\) for targeted group identification), and \(s_j = \tilde{\mathbf{Z}}_{\text{test}} \cdot \tilde{\mathbf{Z}}_j \in [-1, 1]\) is the cosine similarity between the test and neighbor embeddings.

\subsubsection{Similarity-Weighted Label Aggregation}

A soft label distribution is computed from the retrieved neighbors’ labels, weighted by their cosine similarities:

\begin{equation}
	\hat{\mathbf{y}}[c] = \frac{\sum_{j=1}^{k} \mathbb{1}[y_j = c] \cdot s_j}{\sum_{j=1}^{k} s_j}, \quad \forall c \in \{0, \ldots, C-1\},
	\label{eq:faiss-weighted}
\end{equation}
where \(\mathbb{1}[y_j = c]\) is an indicator function equal to 1 if \(y_j = c\) and 0 otherwise, \(s_j\) is the cosine similarity from Equation~\eqref{eq:knn_retrieval}, and \(\hat{\mathbf{y}}[c]\) represents the normalized probability for class \(c\), ensuring \(\sum_{c=0}^{C-1} \hat{\mathbf{y}}[c] = 1\).

The predicted class is determined by selecting the class with the highest aggregated probability:

\begin{equation}
	\hat{y} = \arg\max_c \hat{\mathbf{y}}[c], \quad c \in \{0, 1, \ldots, C-1\},
	\label{eq:knn_vote}
\end{equation}
where \(\hat{\mathbf{y}}[c]\) is the probability for class \(c\), and \(\hat{y}\) is the predicted label.

This approach exploits semantic proximity to assign labels, enhancing robustness for underrepresented classes without requiring model retraining.

\subsection{RAG-Fused DORA (RAG-Inference Fusion)}
\label{subsec:method-rag-xdora}

\texttt{RAG-Fused DORA} enhances xDORA's prediction robustness by incorporating contextual knowledge from semantically similar training samples. The pipeline includes:

\begin{enumerate}
	\item \textbf{Prediction Generation}: The xDORA framework encodes each test sample into a fused multimodal representation \(\mathbf{Z}_{\text{test}}\), which is passed through a classification head to produce logits \(\mathbf{O}_{\text{xdora}}\). These logits are transformed into class probabilities \(\hat{\mathbf{y}}_{\text{xdora}}\) via a softmax (or sigmoid) function, as defined in Equation~\ref{eq:xdora-proba}.
	\item \textbf{Similarity-Weighted Label Distribution Score}: Top-\(k\) nearest neighbors retrieved via FAISS form a soft label distribution, \(\hat{\mathbf{y}}_{\text{faiss}}\) (Equation \ref{eq:faiss-weighted}).
	\item \textbf{Ensemble Fusion}: Final predictions \(\hat{\mathbf{y}}_{\text{fused}}\) combine model-based and retrieval-informed distributions:
	\begin{equation}
		\hat{\mathbf{y}}_{\text{fused}} = \alpha \cdot \hat{\mathbf{y}}_{\text{xdora}} + (1 - \alpha) \cdot \hat{\mathbf{y}}_{\text{faiss}}, \quad \alpha \in [0, 1],
		\label{eq:fused_pred}
	\end{equation}
	with the predicted class as:
	\begin{equation}
		\hat{y} = \arg\max_c \hat{\mathbf{y}}_{\text{fused}}[c].
		\label{eq:fused_class}
	\end{equation}
\end{enumerate}

This approach synthesizes learned representations with non-parametric evidence, improving generalization for ambiguous or underrepresented samples.

\begin{figure*}[!t]
	\centering
	\includegraphics[width=\linewidth]{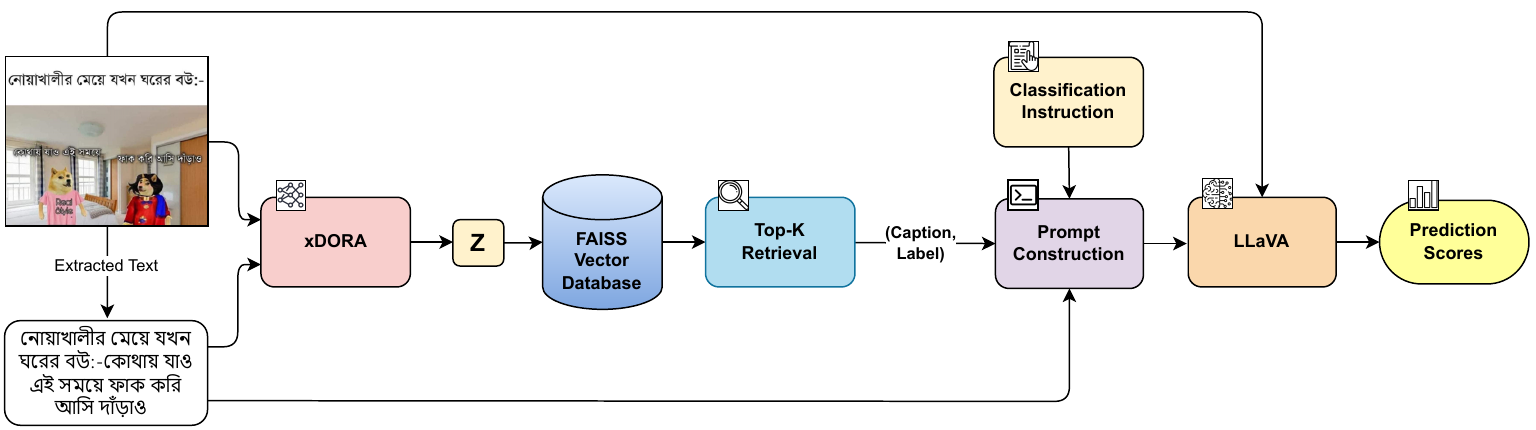}
	\caption{Illustration of the RAG-Prompted LLaVA framework. Given a test image, top-\(k\) semantically similar exemplars per class are retrieved from a FAISS index using xDORA-generated embeddings. The retrieved text-label pairs are formatted as few-shot exemplars and incorporated into a prompt alongside the test caption. This retrieval-augmented prompt is then fed to LLaVA, enabling few-shot multimodal classification.}
	\label{fig:rag-prompted-llava}
\end{figure*}

\begin{algorithm}[htb]	
	\caption{Core Logic of the Multimodal Classification Strategy}
	\label{alg:prompt-structure}
	\KwIn{Image $I$, Caption $C$, Task $T$}
	\KwOut{Predicted label $\hat{y}$}

	\textbf{Step 1: Multimodal Analysis}\;
	\Indp
	Analyze image $I$ for visual cues (e.g., symbols, gestures, expressions)\;
	Analyze caption $C$ for textual signals (e.g., slang, sarcasm, offensive tone)\;
	Perform joint reasoning over $I$ and $C$ to contextualize intent\;
	\Indm

	\textbf{Step 2: Few-Shot Conditioning}\;
	\Indp
	Retrieve top-\(k\) similar examples $\{(C_i, y_i)\}_{i=1}^{k}$ as contextual guidance using Equation \eqref{eq:knn_retrieval}\;
	Learn from example mappings $(C_i \rightarrow y_i)$ to enhance inference\;
	\Indm

	\textbf{Step 3: Classification Based on Task}\;
	\If{$T$ == ``Task 1''}{
		\If{content directly or implicitly targets individuals/groups with hateful intent}{
			$\hat{y} \leftarrow 1$
		}
		\Else{
			$\hat{y} \leftarrow 0$
		}
	}
	\Else{
		Identify the hate target entity according to definitions given in \textbf{BHM} Dataset\;
		\Switch{target type}{
			\Case{TI}{ $\hat{y} \leftarrow 0$ }
			\Case{TC}{ $\hat{y} \leftarrow 1$ }
			\Case{TO}{ $\hat{y} \leftarrow 2$ }
			\Case{TS}{ $\hat{y} \leftarrow 3$ }
		}
	}

	\textbf{Step 4: Ambiguity Resolution}\;
	\Indp
	If ambiguity exists, apply contextual reasoning to select the most plausible label\;
	\Indm

	\Return $\hat{y}$
\end{algorithm}

\subsection{RAG-Prompted LLaVA (RAG-Prompt Fusion)}
\label{subsec:method-rag-llava}

\texttt{RAG-Prompted LLaVA} enhances the few-shot inference of \texttt{LLaVA-1.6-Mistral-7B} by incorporating semantically relevant exemplars into multimodal prompts, leveraging the FAISS-based retriever for context-aware classification. The pipeline includes:

\begin{enumerate}
	\item \textbf{Crossmodal Embedding Retrieval}: For a test instance, the top-\(k\) most semantically similar training samples per class are retrieved using xDORA-generated embeddings, as described in Section \ref{subsec:method-knn}.
	\item \textbf{Prompt Construction}: The captions and labels of the retrieved samples are formatted as text-only few-shot exemplars, ensuring balanced representation across hate categories. The test meme's caption is appended to the prompt as the query. The prompt follows the structure in Algorithm \ref{alg:prompt-structure}, but includes retrieved exemplars to provide context instead of randomly selected examples.
	\item \textbf{Multimodal Inference}: The prompt, comprising the few-shot exemplars and the test caption, is processed alongside the test meme's image by \texttt{LLaVA-1.6-Mistral-7B}. The model leverages the contextual cues from the exemplars to predict whether the meme is hateful and, if applicable, its specific hate category.
\end{enumerate}

By incorporating semantically aligned exemplars, this approach enhances the model's ability to infer nuanced patterns in low-resource settings without requiring parameter updates.

\section{Experiments and Results}
\label{sec:experiments}

This section evaluates the proposed methodologies for multimodal classification of hateful memes in Bengali using the extended Bengali Hateful Memes (BHM) dataset. The methodologies, including the Enhanced Dual cO-attention fRAmework (xDORA), FAISS-Based k-Nearest Neighbor (k-NN) classifier, RAG-Fused DORA, and RAG-Prompted LLaVA, are detailed in Section~\ref{sec:methodology}. The experiments assess performance for hateful meme identification (Task 1: binary classification; hateful vs. non-hateful) and target entity detection (Task 2: multiclass classification; TI, TC, TO, TS), focusing on hardware configurations, hyperparameter settings, evaluation metrics, baseline models, quantitative and qualitative analyses, and comparative insights against state-of-the-art unimodal and multimodal models.

\subsection{Experimental Setup}
\label{subsec:experimental_setup}


Experiments were conducted on Google Colab and Kaggle using NVIDIA T4 or L4 GPUs (16 GB VRAM). Automatic Mixed Precision (AMP) was employed to optimize memory and computation efficiency.


The xDORA models were trained for Task 1 (binary classification: hateful vs. non-hateful) and Task 2 (multiclass classification: TI, TC, TO, TS) using four encoder pairs: CLIP (ViT-B/32)~\cite{radford2021learning} or DINOv2-Base~\cite{oquab2023dinov2} with XGLM-564M~\cite{xglm} or XLM-R-L~\cite{xlmroberta}. Training used weighted CrossEntropyLoss, batch size of 16, learning rate of \(2 \times 10^{-5}\), and weight decay of 0.01. Models with XGLM-564M used MADGRAD~\cite{defazio2021adaptivity}, while those with XLM-R-L used AdamW~\cite{loshchilov2017decoupled}, with up to 20 epochs and early stopping based on validation accuracy. Dropout (0.1) was applied in the MLP layer.

LLaVA-1.6-Mistral-7B~\cite{llava1.6mistral} in RAG-Prompted settings used 4-bit quantization. Prompts included five FAISS-retrieved caption-label exemplars per class for contextual supervision, as per Algorithm~\ref{alg:prompt-structure}. Zero-shot and few-shot settings used five randomly sampled caption-label pairs per class, as detailed in Section~\ref{subsec:method-rag-llava}.

The FAISS-based k-NN classifier, RAG-Fused DORA, and RAG-Prompted LLaVA used xDORA embeddings (CLIP + XLM-R-L, selected for F1-scores of 0.78 and 0.71 for Tasks 1 and 2, Table~\ref{tab:performance-stats}). Embeddings (\(\mathbf{Z} \in \mathbb{R}^{4096}\), Equation~\ref{eq:fused_output}) were \(\ell_2\)-normalized and indexed in a FAISS flat index with cosine similarity (Section~\ref{subsec:method-knn}). The k-NN classifier used \(k=5\) neighbors for majority voting. RAG-Fused DORA retrieved \(k=5\) neighbors per class, interpolating xDORA logits with similarity-weighted label distributions (\(\alpha=0.6\), grid-searched over [0.5, 0.7]). RAG-Prompted LLaVA used \(k=5\) text-only exemplars per class for prompt augmentation (Section~\ref{subsec:method-rag-llava}).

\subsubsection{Evaluation Metrics}

Performance was evaluated using macro-averaged precision, recall, and F1-score to address class imbalance, particularly in Task 2. Confidence intervals were computed via 1000-iteration bootstrapping, ensuring robust evaluation for underrepresented classes like TS, which are prone to performance variability.

\begin{table*}[!htb]
    \centering
    \caption{Performance comparison of baseline and proposed models on the test set using macro-average metrics. Models are categorized as Vision Only, Text Only, Multimodal, LVLM, Proposed Supervised Framework Variants, and Proposed Retrieval Augmented Methods. Metrics include macro-average precision (P), recall (R), and F1-score (F1) for Task 1 (hateful meme identification) and Task 2 (target entity detection), reported as mean \(\pm\) {\scriptsize 95\% confidence interval half-width}. Highest and second-highest F1-scores are in \textbf{bold} and \underline{underlined}, respectively.}
    \label{tab:performance-stats}
	\resizebox{\linewidth}{!}{%
		\begin{tabular}{>{\footnotesize\raggedright\arraybackslash}p{2cm}|>{\small\raggedright\arraybackslash}p{3.3cm}|ccc|ccc}
			\toprule
			\multirow{2}{*}{\textbf{Category}} & \multirow{2}{*}{\textbf{Model}} & \multicolumn{3}{c|}{\textbf{Hateful Meme Identification (Task 1)}} & \multicolumn{3}{c}{\textbf{Target Entity Detection (Task 2)}} \\
			\cmidrule{3-5} \cmidrule{6-8}
			& & \textbf{P} & \textbf{R} & \textbf{F1} & \textbf{P} & \textbf{R} & \textbf{F1} \\
			\midrule
            \multirow{3}{*}{Vision Only} 
                & CLIP (ViT-B/32) & 0.50 $\pm$ {\scriptsize 0.018} & 0.50 $\pm$ {\scriptsize 0.015} & 0.48 $\pm$ {\scriptsize 0.015} 
                                          & 0.41 $\pm$ {\scriptsize 0.015} & 0.44 $\pm$ {\scriptsize 0.018} & 0.42 $\pm$ {\scriptsize 0.015} \\
            & DINOv2-Base & 0.65 $\pm$ {\scriptsize 0.018} & 0.65 $\pm$ {\scriptsize 0.015} & 0.65 $\pm$ {\scriptsize 0.018} 
                           & 0.48 $\pm$ {\scriptsize 0.023} & 0.51 $\pm$ {\scriptsize 0.038} & 0.48 $\pm$ {\scriptsize 0.025} \\
            & ConvNeXT-Base & 0.63 $\pm$ {\scriptsize 0.018} & 0.62 $\pm$ {\scriptsize 0.015} & 0.61 $\pm$ {\scriptsize 0.015} 
                            & 0.44 $\pm$ {\scriptsize 0.018} & 0.46 $\pm$ {\scriptsize 0.018} & 0.45 $\pm$ {\scriptsize 0.015} \\
            \midrule
            \multirow{3}{*}{Text Only} 
                & XGLM-564M & 0.75 $\pm$ {\scriptsize 0.013} & 0.74 $\pm$ {\scriptsize 0.015} & 0.74 $\pm$ {\scriptsize 0.015} 
                             & 0.55 $\pm$ {\scriptsize 0.018} & 0.61 $\pm$ {\scriptsize 0.038} & 0.56 $\pm$ {\scriptsize 0.023} \\
            & XLM-R-L & 0.73 $\pm$ {\scriptsize 0.013} & 0.73 $\pm$ {\scriptsize 0.013} & 0.73 $\pm$ {\scriptsize 0.013} 
                                & 0.65 $\pm$ {\scriptsize 0.023} & 0.77 $\pm$ {\scriptsize 0.030} & 0.69 $\pm$ {\scriptsize 0.025} \\
            & mDeBERTa-v3 & 0.74 $\pm$ {\scriptsize 0.013} & 0.74 $\pm$ {\scriptsize 0.013} & 0.74 $\pm$ {\scriptsize 0.013} 
                                & 0.60 $\pm$ {\scriptsize 0.025} & 0.62 $\pm$ {\scriptsize 0.043} & 0.60 $\pm$ {\scriptsize 0.033} \\
            \midrule
            \multirow{2}{*}{Multimodal} 
                & DORA & 0.76 $\pm$ {\scriptsize 0.015} & 0.76 $\pm$ {\scriptsize 0.015} & 0.76 $\pm$ {\scriptsize 0.015} 
                       & 0.61 $\pm$ {\scriptsize 0.013} & 0.59 $\pm$ {\scriptsize 0.015} & 0.60 $\pm$ {\scriptsize 0.015} \\
            & MAF & 0.77 $\pm$ {\scriptsize 0.013} & 0.77 $\pm$ {\scriptsize 0.013} & 0.77 $\pm$ {\scriptsize 0.013} 
                  & 0.56 $\pm$ {\scriptsize 0.015} & 0.55 $\pm$ {\scriptsize 0.015} & 0.56 $\pm$ {\scriptsize 0.015} \\
            \midrule
            \multirow{2}{*}{LVLM} 
                & LLaVA Zero-Shot & 0.55 $\pm$ {\scriptsize 0.018} & 0.54 $\pm$ {\scriptsize 0.015} & 0.54 $\pm$ {\scriptsize 0.015} 
                                   & 0.41 $\pm$ {\scriptsize 0.040} & 0.37 $\pm$ {\scriptsize 0.023} & 0.21 $\pm$ {\scriptsize 0.020} \\
            & LLaVA Few-Shot & 0.53 $\pm$ {\scriptsize 0.015} & 0.53 $\pm$ {\scriptsize 0.015} & 0.53 $\pm$ {\scriptsize 0.015} 
                              & 0.29 $\pm$ {\scriptsize 0.033} & 0.30 $\pm$ {\scriptsize 0.025} & 0.26 $\pm$ {\scriptsize 0.018} \\
            \midrule
            \multirow{4}{*}{\parbox{2cm}{Proposed Supervised Framework Variants}} 
                & CLIP + XGLM & 0.78 $\pm$ {\scriptsize 0.013} & 0.77 $\pm$ {\scriptsize 0.013} & 0.78 $\pm$ {\scriptsize 0.013} 
                                  & 0.69 $\pm$ {\scriptsize 0.048} & 0.66 $\pm$ {\scriptsize 0.038} & 0.67 $\pm$ {\scriptsize 0.040} \\
            & CLIP + XLM-R-L & \underline{0.79 $\pm$ {\scriptsize 0.013}} & \underline{0.78 $\pm$ {\scriptsize 0.013}} & \underline{0.78 $\pm$ {\scriptsize 0.013}} 
                             & \underline{0.70 $\pm$ {\scriptsize 0.033}} & \underline{0.70 $\pm$ {\scriptsize 0.040}} & \underline{0.71 $\pm$ {\scriptsize 0.033}} \\
            & DINOv2 + XGLM & 0.76 $\pm$ {\scriptsize 0.013} & 0.76 $\pm$ {\scriptsize 0.013} & 0.76 $\pm$ {\scriptsize 0.015} 
                                      & 0.66 $\pm$ {\scriptsize 0.018} & 0.78 $\pm$ {\scriptsize 0.033} & 0.67 $\pm$ {\scriptsize 0.023} \\
            & DINOv2 + XLM-R & 0.77 $\pm$ {\scriptsize 0.015} & 0.77 $\pm$ {\scriptsize 0.015} & 0.77 $\pm$ {\scriptsize 0.015} 
                                    & 0.67 $\pm$ {\scriptsize 0.020} & 0.78 $\pm$ {\scriptsize 0.033} & 0.68 $\pm$ {\scriptsize 0.023} \\
            \midrule
            \multirow{3}{*}{\parbox{2.25cm}{Proposed Retrieval-Based Methods}} 
                & FAISS-Based k-NN & 0.78 $\pm$ {\scriptsize 0.013} & 0.77 $\pm$ {\scriptsize 0.013} & 0.77 $\pm$ {\scriptsize 0.013} 
                              & 0.76 $\pm$ {\scriptsize 0.043} & 0.71 $\pm$ {\scriptsize 0.040} & 0.73 $\pm$ {\scriptsize 0.038} \\
            & RAG-Fused DORA & \textbf{0.79 $\pm$ {\scriptsize 0.013}} & \textbf{0.79 $\pm$ {\scriptsize 0.013}} & \textbf{0.79 $\pm$ {\scriptsize 0.013}} 
                             & \textbf{0.74 $\pm$ {\scriptsize 0.033}} & \textbf{0.74 $\pm$ {\scriptsize 0.040}} & \textbf{0.74 $\pm$ {\scriptsize 0.033}} \\
            & RAG-Prompted LLaVA & 0.55 $\pm$ {\scriptsize 0.015} & 0.55 $\pm$ {\scriptsize 0.015} & 0.54 $\pm$ {\scriptsize 0.015} 
                                        & 0.48 $\pm$ {\scriptsize 0.030} & 0.42 $\pm$ {\scriptsize 0.035} & 0.39 $\pm$ {\scriptsize 0.025} \\
            \bottomrule
        \end{tabular}
    }
\end{table*}

\subsubsection{Baseline Models}
\label{subsubsec:baseline_models}

Baseline models span Vision Only, Text Only, Multimodal, and Large Vision-Language Model (LVLM) categories, fine-tuned on the extended BHM dataset using the training protocols outlined in Subsection~\ref{subsec:experimental_setup} (batch size 16, learning rate \(2 \times 10^{-5}\), up to 20 epochs with early stopping), except for LLaVA Zero-Shot and LLaVA Few-Shot, which do not undergo fine-tuning. 

\paragraph{Vision Only Baselines}

\begin{itemize}
    \item \textbf{CLIP (ViT-B/32)}: Generates 512-dimensional embeddings with a frozen backbone and fine-tuned fully connected layers.
    \item \textbf{DINOv2-Base}: Yields 768-dimensional embeddings with a frozen backbone and fine-tuned fully connected layers.
    \item \textbf{ConvNeXT-Base}: Produces 1024-dimensional embeddings using depth-wise convolutions and large kernel sizes \cite{liu2022convnet}.
\end{itemize}

\paragraph{Text Only Baselines}

\begin{itemize}
    \item \textbf{XGLM-564M}: Generates 1024-dimensional embeddings, fine-tuned with MADGRAD.
    \item \textbf{XLM-R-L}: Produces 1024-dimensional embeddings, fine-tuned with AdamW.
    \item \textbf{mDeBERTa-v3-base}: Yields 768-dimensional embeddings optimized for low-resource languages \cite{he2021deberta}.
\end{itemize}

\paragraph{Multimodal Baselines}

\begin{itemize}
    \item \textbf{DORA}: Integrates CLIP (ViT-B/32) and XGLM-564M with a Dual Co-Attention module \cite{hossain2024deciphering}.
    \item \textbf{MAF}: Combines frozen CLIP (ViT-B/32) and Bangla-BERT-Base with multi-head attention \cite{ahsan2024multimodal}.
\end{itemize}

\paragraph{LVLM Baselines}

\begin{itemize}
    \item \textbf{LLaVA Zero-Shot}: Uses LLaVA-1.6-Mistral-7B, quantized to 4-bit, for inference without fine-tuning.
    \item \textbf{LLaVA Few-Shot}: Employs LLaVA-1.6-Mistral-7B with five caption-label exemplars per class in the prompt, without parameter updates.
\end{itemize}

\subsection{Results and Analysis}
\label{subsec:results}

\begin{table}[!htbp]
    \centering
    \caption{Class-wise performance of xDORA (\texttt{CLIP + XLM-R-L}), FAISS-Based k-NN, and RAG-Fused DORA for Task 2. Metrics include precision (P), recall (R), F1-score (F1), and support. Highest F1-score per class is in \textbf{bold}.}
    \label{tab:classwise-performance}
    \small
    \resizebox{.9\columnwidth}{!}{%
        \begin{tabular}{@{}l|c|cccc@{}}
            \toprule
            Model & Class & \textbf{P} & \textbf{R} & \textbf{F1} & \textbf{Support} \\
            \midrule
            \multirow{4}{*}{CLIP + XLM-R-L} 
                & TI & 0.83 & 0.87 & \textbf{0.85} & 254 \\
                & TC & 0.76 & 0.76 & 0.76 & 122 \\
                & TO & 0.84 & 0.71 & \textbf{0.77} & 99 \\
                & TS & 0.38 & 0.55 & 0.44 & 11 \\
            \midrule
            \multirow{4}{*}{FAISS-Based k-NN} 
                & TI & 0.80 & 0.89 & 0.84 & 254 \\
                & TC & 0.81 & 0.70 & 0.75 & 122 \\
                & TO & 0.77 & 0.69 & 0.73 & 99 \\
                & TS & 0.67 & 0.55 & \textbf{0.60} & 11 \\
            \midrule
            \multirow{4}{*}{RAG-Fused DORA} 
                & TI & 0.82 & 0.89 & \textbf{0.85} & 254 \\
                & TC & 0.80 & 0.74 & \textbf{0.77} & 122 \\
                & TO & 0.83 & 0.71 & \textbf{0.77} & 99 \\
                & TS & 0.50 & 0.64 & 0.56 & 11 \\
            \bottomrule
        \end{tabular}
    }
\end{table}

\begin{table*}[!ht]
    \centering
	\caption{Sample evaluations of memes from the test set for Task 2 (multiclass classification). The table includes the meme image, OCR-extracted caption, translations, predictions from baseline models, proposed framework variants, and the ground truth. Correct predictions are shown in \textcolor{ForestGreen}{green} and incorrect predictions in \textcolor{BrickRed}{red}.}    
	\label{tab:eval-task2}
    \small
    \setlength{\tabcolsep}{2pt}
    \resizebox{.9\linewidth}{!}{%
        \begin{tabular}{@{}l|c|c|c|c@{}}
            & (a) & (b) & (c) & (d) \\
            \toprule
            \textbf{Meme} 
            &
            \includegraphics[width=.13\linewidth]{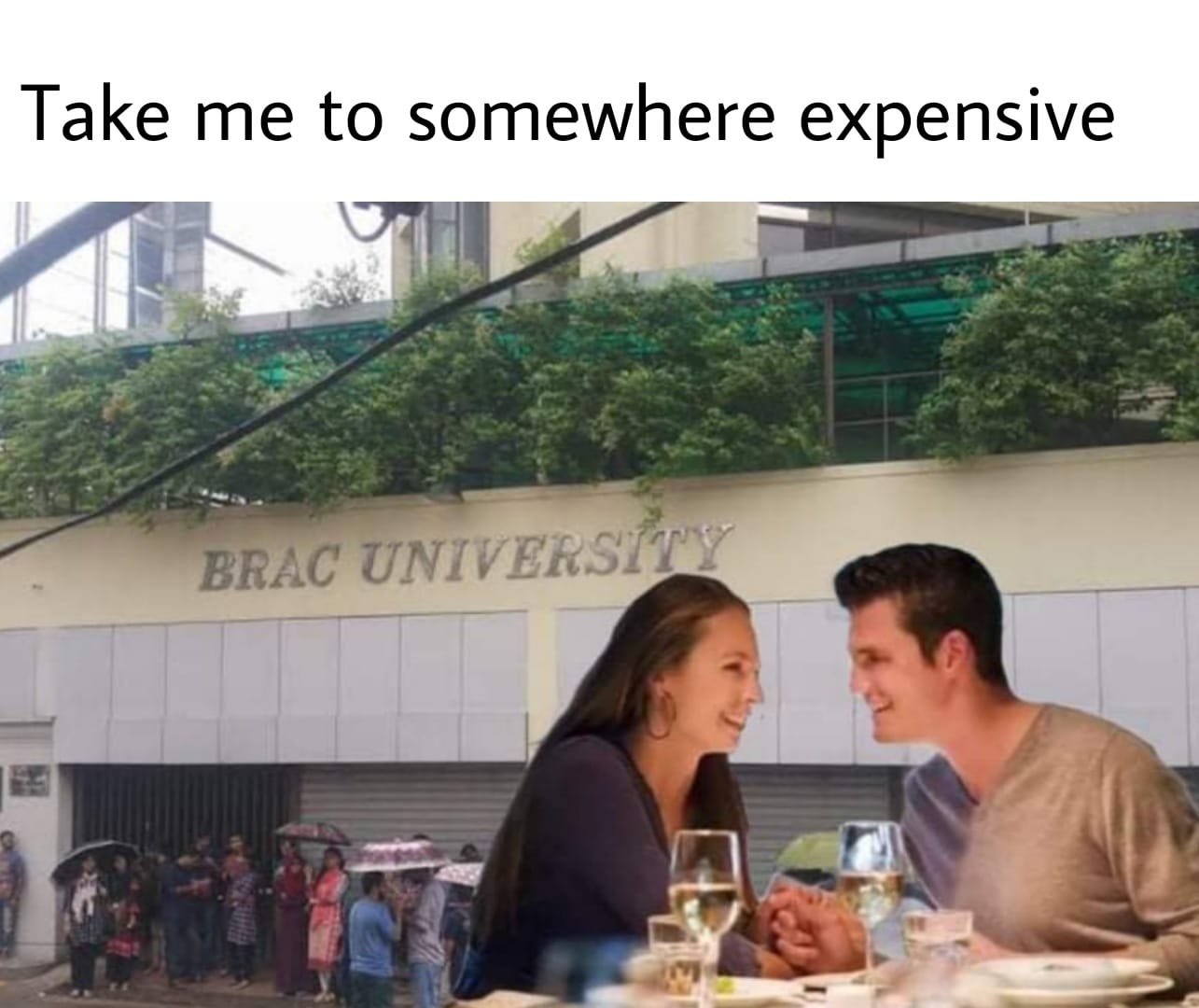} &
            \includegraphics[width=0.13\linewidth]{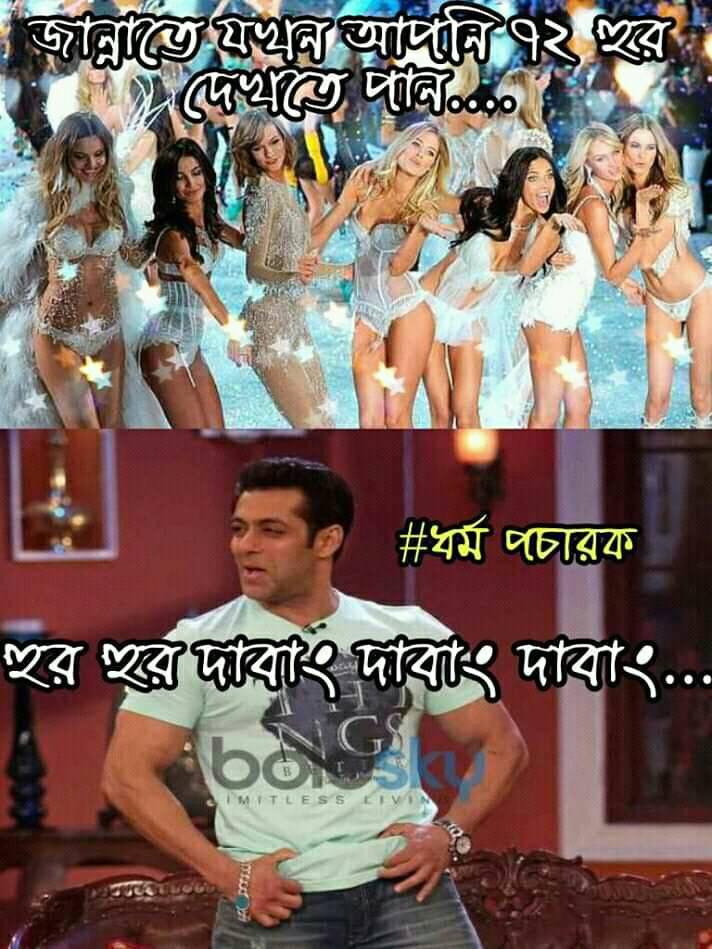} &
            \includegraphics[width=0.13\linewidth]{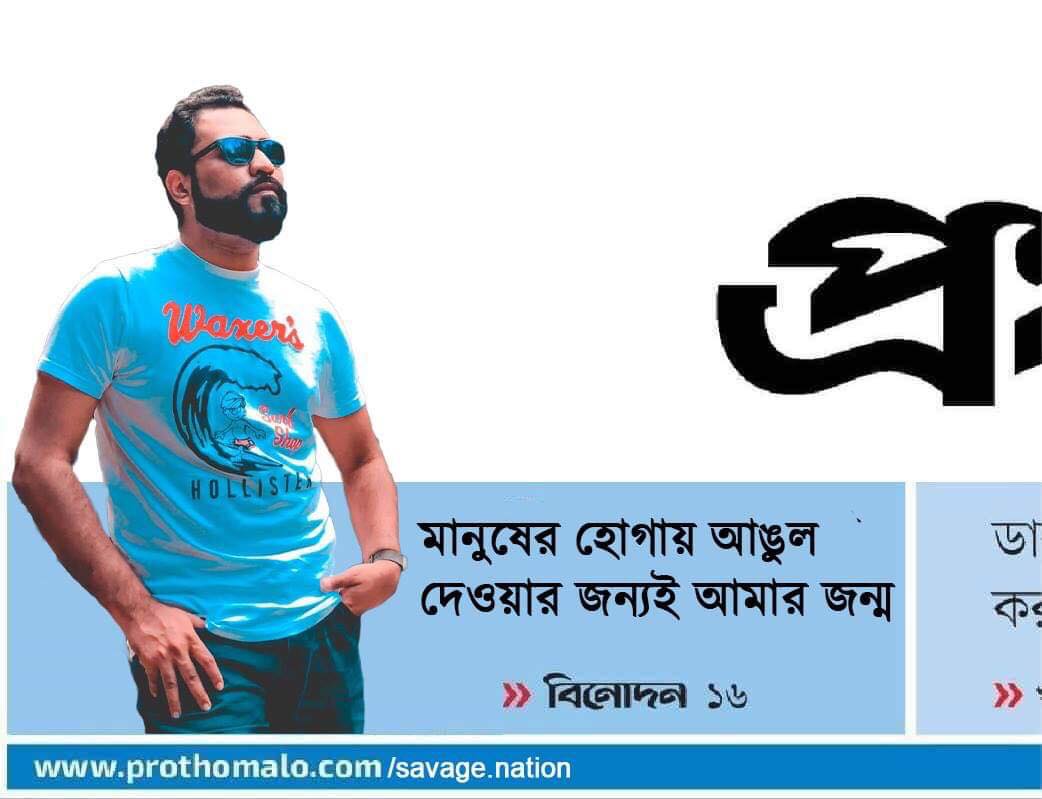} &
            \includegraphics[width=0.13\linewidth]{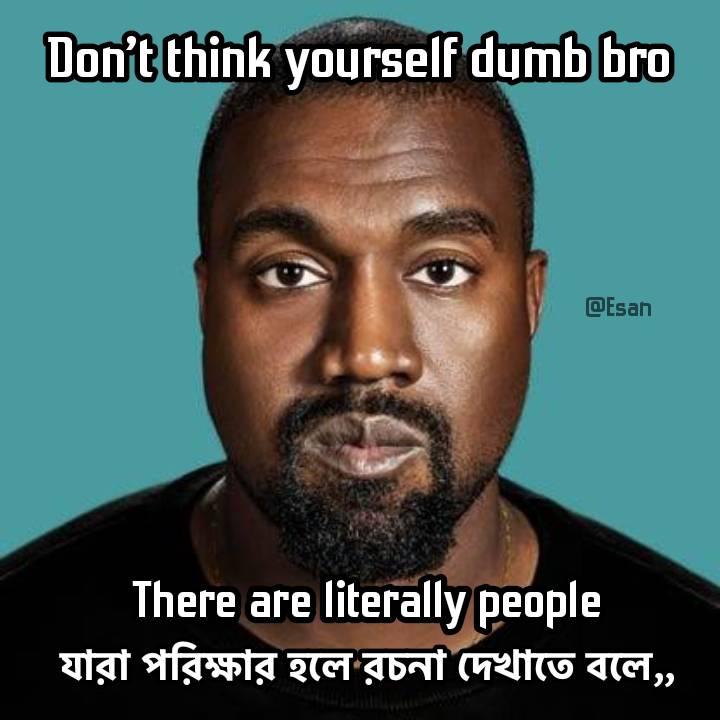} \\
            \midrule
            \textbf{Caption} &
            \footnotesize\begin{varwidth}{2.7cm} Take me to somewhere expensive BRAC UNIVERSITY \end{varwidth} &
            \footnotesize\begin{varwidth}{2.7cm} {\bangla জন্নাতে যখন আপনি ৭২ হুর দেখতে পান... \#ধর্ম পচারক হুর হুর দাবাং দাবাং দাবাং...} \end{varwidth} &
            \footnotesize\begin{varwidth}{2.7cm} {\bangla মানুষের হো*য় আঙুল দেওয়ার জন্যই আমার জন্ম} \end{varwidth} &
            \footnotesize\begin{varwidth}{2.7cm} {\bangla Don't think yourself dumb bro There are literally people যারা পরিক্ষার হলে রচনা দেখাতে বলে ,,} \end{varwidth} \\
            \midrule
            \textbf{Translated Caption} &
            \footnotesize\begin{varwidth}{2.7cm} Take me to somewhere expensive BRAC UNIVERSITY \end{varwidth} &
            \footnotesize\begin{varwidth}{2.7cm} When you see 72 houris in heaven… \#ReligiousPreacher Hoor hoor dabangg dabangg dabangg… \end{varwidth} &
            \footnotesize\begin{varwidth}{2.7cm} I was born just to poke fingers at people's a*se. \end{varwidth} &
            \footnotesize\begin{varwidth}{2.7cm} Don't think yourself dumb, bro. There are literally people who ask to copy essays during exams… \end{varwidth} \\
            \midrule
            \textbf{Vision Only} & \textcolor{BrickRed}{TI} & \textcolor{BrickRed}{TI} & \textcolor{BrickRed}{TC} & \textcolor{BrickRed}{TI} \\
            \textbf{Text Only} & \textcolor{BrickRed}{TS} & \textcolor{ForestGreen}{TC} & \textcolor{BrickRed}{TC} & \textcolor{BrickRed}{TI} \\
            \textbf{Multimodal} & \textcolor{ForestGreen}{TO} & \textcolor{ForestGreen}{TC} & \textcolor{ForestGreen}{TI} & \textcolor{BrickRed}{TI} \\
			\textbf{LLaVA (Zero-Shot)} & \textcolor{BrickRed}{TC} & \textcolor{BrickRed}{TO} & \textcolor{BrickRed}{TC} & \textcolor{BrickRed}{TO} \\
            \textbf{LLaVA (Few-Shot)} & \textcolor{BrickRed}{TC} & \textcolor{BrickRed}{TO} & \textcolor{BrickRed}{TC} & \textcolor{BrickRed}{TO} \\
			\midrule
            \textbf{CLIP + XGLM} & \textcolor{BrickRed}{TC} & \textcolor{BrickRed}{TS} & \textcolor{BrickRed}{TC} & \textcolor{BrickRed}{TC} \\
            \textbf{CLIP + XLM-R-L} & \textcolor{ForestGreen}{TO} & \textcolor{ForestGreen}{TC} & \textcolor{ForestGreen}{TI} & \textcolor{BrickRed}{TI} \\
            \textbf{DINOv2 + XGLM} & \textcolor{BrickRed}{TS} & \textcolor{BrickRed}{TS} & \textcolor{ForestGreen}{TI} & \textcolor{BrickRed}{TC} \\
            \textbf{DINOv2 + XLM-R-L} & \textcolor{ForestGreen}{TO} & \textcolor{ForestGreen}{TC} & \textcolor{ForestGreen}{TI} & \textcolor{BrickRed}{TI} \\
            \textbf{FAISS-Based k-NN} & \textcolor{ForestGreen}{TO} & \textcolor{ForestGreen}{TC} & \textcolor{ForestGreen}{TI} & \textcolor{ForestGreen}{TC} \\
            \textbf{RAG-Fused DORA} & \textcolor{ForestGreen}{TO} & \textcolor{ForestGreen}{TC} & \textcolor{ForestGreen}{TI} & \textcolor{ForestGreen}{TC} \\
            \textbf{RAG-Prompted LLaVA} & \textcolor{BrickRed}{TC} & \textcolor{ForestGreen}{TC} & \textcolor{BrickRed}{TC} & \textcolor{BrickRed}{TO} \\
            \midrule
            \textbf{Ground Truth} & TO & TC & TI & TC \\
            \bottomrule
        \end{tabular}
    }
\end{table*}

This subsection presents the performance of the proposed methodologies—Enhanced Dual cO-attention fRAmework (xDORA), FAISS-Based k-Nearest Neighbor (k-NN) classifier, RAG-Fused DORA, and RAG-Prompted LLaVA—on the extended Bengali Hateful Memes (BHM) dataset, alongside baseline models. The evaluation focuses on two tasks: hateful meme identification (Task 1: binary classification, hateful vs. non-hateful) and target entity detection (Task 2: multiclass classification, TI, TC, TO, TS). Results are reported using macro-averaged precision, recall, and F1-scores, with confidence intervals derived from 1000 bootstrap iterations, as detailed in Table~\ref{tab:performance-stats}. Qualitative insights from sample evaluations are provided in Table~\ref{tab:eval-task2}.

\subsubsection{Hateful Meme Detection}
\label{subsubsec:hateful_meme_detection}

Task 1 evaluates the ability of models to distinguish hateful from non-hateful memes. RAG-Fused DORA achieves the highest macro-averaged F1-score of 0.79 $\pm$ 0.013, demonstrating superior contextual sensitivity through its integration of xDORA’s predictions with similarity-weighted label distributions from retrieved samples. The xDORA model (CLIP + XLM-R-L) follows closely with an F1-score of 0.78 $\pm$ 0.013, leveraging enhanced multimodal fusion and cross-lingual processing. Other xDORA variants (CLIP + XGLM, DINOv2 + XGLM, DINOv2 + XLM-R-L) yield F1-scores between 0.76 and 0.78, with confidence intervals of $\pm 0.013–0.015$, indicating reliable performance. The FAISS-Based k-NN classifier performs competitively with an F1-score of 0.77 $\pm$ 0.013, benefiting from semantic proximity in the embedding space.

Baseline models exhibit lower performance. Vision-only models (CLIP: 0.48, DINOv2-Base: 0.65, ConvNeXT-Base: 0.61) struggle due to their reliance on visual features alone. Text-only models (XGLM-564M: 0.74, XLM-R-L: 0.73, mDeBERTa-v3-base: 0.74) perform better but are limited by the absence of visual context. Multimodal baselines (DORA: 0.76, MAF: 0.77) approach the proposed methods but have wider confidence intervals ($\pm 0.015$), indicating less stability. Large Vision-Language Model (LVLM) baselines (LLaVA Zero-Shot: 0.54, LLaVA Few-Shot: 0.53) underperform due to insufficient fine-tuning for Bengali memes, with RAG-Prompted LLaVA slightly improving to 0.54 $\pm$ 0.015 through retrieval augmentation.

\subsubsection{Target Entity Identification}
\label{subsubsec:target_entity_identification}

Task 2 involves classifying memes into specific hate target categories: Targeted Individual (TI), Targeted Community (TC), Targeted Organization (TO), and Targeted Social Group (TS). Table~\ref{tab:performance-stats} shows RAG-Fused DORA leading with an F1-score of 0.74 $\pm$ 0.033, followed by FAISS-Based k-NN at 0.73 $\pm$ 0.038 and xDORA (CLIP + XLM-R-L) at 0.71 $\pm$ 0.033. These models excel due to their ability to integrate visual and textual cues effectively, with RAG-Fused DORA benefiting from retrieval-augmented supervision. Other xDORA variants achieve F1-scores of 0.67–0.68, with confidence intervals of $\pm 0.023–0.040$, reflecting consistent performance despite class imbalance (e.g., 254 TI samples vs. 11 TS samples).

Class-wise performance in Table~\ref{tab:classwise-performance} highlights the models’ strengths. For TI (254 samples), xDORA (CLIP + XLM-R-L) and RAG-Fused DORA achieve the highest F1-score of 0.85, while FAISS-Based k-NN scores 0.84. For TC (122 samples), RAG-Fused DORA leads with 0.77, followed by xDORA at 0.76. For TO (99 samples), xDORA and RAG-Fused DORA both score 0.77. For the underrepresented TS class (11 samples), FAISS-Based k-NN achieves the highest F1-score of 0.60, leveraging semantic proximity to address data scarcity, while RAG-Fused DORA and xDORA score 0.56 and 0.44, respectively.

Baseline models lag behind. Vision-only models (CLIP: 0.42, DINOv2-Base: 0.48, ConvNeXT-Base: 0.45) and text-only models (XGLM-564M: 0.56, XLM-R-L: 0.69, mDeBERTa-v3-base: 0.60) struggle with multimodal nuances. Multimodal baselines (DORA: 0.60, MAF: 0.56) perform better but are outperformed by the proposed methods. LVLM baselines (LLaVA Zero-Shot: 0.21, LLaVA Few-Shot: 0.26) show poor performance, with RAG-Prompted LLaVA improving to 0.39 $\pm$ 0.025 but remaining limited by English-centric pretraining.

\subsubsection{Error Analysis}
\label{subsubsec:error_analysis}

To gain insights into the model’s mistakes, we conduct a qualitative error analysis by examining some correctly and incorrectly classified samples, as illustrated in Table~\ref{tab:eval-task2}. Classification errors in Task 2, as shown in Table~\ref{tab:eval-task2}, arise from overlapping hate category definitions, scarcity of TS instances (11 samples), and cultural nuances in code-mixed captions, reflected in the moderate inter-annotator agreement ($\kappa = 0.63$). Sample evaluations illustrate specific challenges. For a meme critiquing BRAC University’s high costs (TO), Multimodal (DORA), xDORA (CLIP + XLM-R-L, DINOv2 + XLM-R-L), FAISS-Based k-NN, and RAG-Fused DORA correctly predict TO due to effective multimodal integration, while Vision Only (DINOv2-Base), Text Only (XLM-R-L), and LLaVA variants misclassify as TI, TS, or TC due to reliance on isolated modalities. A meme mocking religious promises with phrases like “{\bangla ৭২ হুর}” (72 houris) (TC) is accurately identified by Text Only, Multimodal, xDORA, FAISS-Based k-NN, and RAG-Fused DORA, but Vision Only and LLaVA models misclassify as TI or TO, misinterpreting religious imagery due to limited Bengali fine-tuning. A meme using vulgar humor (“{\bangla মানুষের হো*য় আঙুল দেওয়ার জন্যই}”) to target individuals (TI) is correctly classified by Multimodal, xDORA, FAISS-Based k-NN, and RAG-Fused DORA, while Vision Only, Text Only, and LLaVA variants misclassify as TC, struggling with idiomatic expressions. A meme critiquing exam cheating (TC) is accurately predicted by FAISS-Based k-NN and RAG-Fused DORA, but others misclassify as TI or TC due to ambiguous individual-focused cues or cultural nuances.

Vision-only models frequently fail by relying solely on visual features, missing textual context. Text-only models struggle with idiomatic or culturally specific phrases. LLaVA models (Zero-Shot, Few-Shot, RAG-Prompted) consistently underperform due to English-centric pretraining. The proposed models mitigate these issues through robust multimodal reasoning and retrieval augmentation, though errors persist in cases with ambiguous cues, such as meme~\ref{tab:eval-task2}(d). FAISS-Based k-NN and RAG-Fused DORA excel in addressing class imbalance, particularly for TS, by leveraging semantic proximity and contextual supervision, respectively.

\section{Conclusion}
\label{sec:conclusion}

This study advances hate speech detection in Bengali by evaluating multimodal models on an extended Bengali Hateful Memes (BHM) dataset, designed to address class imbalance. The proposed Enhanced Dual cO-attention fRAmework (xDORA), particularly the \texttt{CLIP + XLM-RoBERTa-Large} configuration, significantly outperforms baseline models, in both hateful meme identification (Task 1) and target entity detection (Task 2). Its effectiveness derives from robust multimodal fusion, integrating fine-tuned cross-lingual text processing with visual feature extraction, making it highly suitable for low-resource language contexts.

The RAG-Fused DORA model achieves the highest performance across both tasks by combining FAISS-based retrieval with supervised learning, enhancing contextual accuracy, particularly for underrepresented classes. The FAISS-Based k-Nearest Neighbor (k-NN) classifier, leveraging xDORA embeddings, provides a competitive non-parametric approach, effectively mitigating class imbalance through semantic proximity. The \texttt{DINOv2-Base}-based xDORA variants further excel in capturing nuanced visual cues in culturally complex memes, contributing to their superior performance over baselines.

In contrast, the \texttt{LLaVA-1.6-Mistral-7B} model underperforms in zero-shot and few-shot settings due to limited fine-tuning and challenges with Bengali’s linguistic and cultural nuances. The RAG-Prompted LLaVA improves slightly by incorporating retrieved exemplars but remains limited by its English-centric pretraining.

Future work includes fine-tuning advanced vision-language models, such as next-generation LLaVA or multimodal architectures like GPT-4o, on Bengali datasets to better capture code-mixed and culturally specific patterns. Extending these methods to related tasks, such as sentiment analysis or toxicity detection, could enhance their applicability in Bengali social media moderation. Additionally, leveraging more powerful computational resources (e.g., NVIDIA A100 GPUs) and distributed optimization strategies would enable exploration of larger architectures and richer multimodal pretraining objectives, overcoming the computational constraints encountered in this study.

\printbibliography

\end{document}